# PESTS: Persian_English Cross Lingual Corpus for Semantic Textual Similarity

Mohammad Abdous [a], Poorya Piroozfar [a] and Behrouz Minaei Bidgoli [a,1]
[a] Iran University of Science and Technology, Department of Computer Engineering, Tehran, Iran
E-mails: mohammadabdous@comp.iust.ac.ir, poorya_piroozfar@alumni.iust.ac.ir, b_minaei@iust.ac.ir

**Abstract**
In recent years, there has been significant research interest in the subtask of natural language processing called semantic textual similarity. Measuring semantic similarity between words or terms, sentences, paragraphs and documents plays an important role in natural language processing and computational linguistics. It finds applications in question-answering systems, semantic search, fraud detection, machine translation, information retrieval, and more. Semantic similarity entails evaluating the extent of similarity in meaning between two textual documents, paragraphs, or sentences, both in the same language and across different languages. To achieve cross-lingual semantic similarity, it is essential to have corpora that consist of sentence pairs in both the source and target languages. These sentence pairs should demonstrate a certain degree of semantic similarity between them. Due to the lack of available cross-lingual semantic similarity datasets, many current models in this field rely on machine translation. However, this dependence on machine translation can result in reduced model accuracy due to the potential propagation of translation errors. In the case of Persian, which is categorized as a low-resource language, there has been a lack of efforts in developing models that can comprehend the context of two languages. The demand for such a model that can bridge the understanding gap between languages is now more crucial than ever. In this article, the corpus of semantic textual similarity between sentences in Persian and English languages has been produced for the first time through the collaboration of linguistic experts. We named this dataset PESTS (**P**ersian **E**nglish **S**emantic **T**extual **S**imilarity). This corpus contains 5375 sentence pairs. Furthermore, various Transformer-based models have been fine-tuned using this dataset. Based on the results obtained from the PESTS dataset, it is observed that the utilization of the XLM_ROBERTa model leads to an increase in the Pearson correlation from 85.87% to 95.62%.

## 1 Introduction

Measuring the similarity in meaning between different sections of text, whether they are individual words, sentences, paragraphs, or entire documents, is a highly significant area of study within the field of natural language processing. The measurement of semantic similarity between sentences plays a crucial role in various natural language applications. These applications include semantic search (Manjula and Geetha, 2004), summarization (Aliguliyev, 2009), question-answering Systems (De Boni and Manandhar, 2003), document classification (Al-Anzi and AbuZeina, 2017), sentiment Analysis (Žižka and Dařena, 2010) and plagiarism detection (Alzahrani *et al.*, 2011). Assessing the degree of semantic similarity with the goal of understanding natural language is attractive research in computer science, artificial intelligence, and computational linguistics. From 2006 to 2012, semantic similarity was framed as a binary classification task (Majumder *et al.*, 2016), involving the determination of whether two sentences possessed semantic similarity. However, since 2012 and continuing to the present day, the quantification of similarity in numerical terms has been adopted, as indicated by (Majumder *et al.*, 2016).

---

[1] Corresponding Author

Sentences in different languages can also be semantically similar. For example, the English sentence "He wants to play football" and the Persian sentence "احمد در ورزشگاه تختی تمرین می‌کند. (English: Ahmed trains in Takhti gym.)" can be semantically similar with score 3 (The two sentences are similar, but there are some key differences. For instance, Ahmed is not mentioned in the second sentence, and there are variations in the activities mentioned, such as trains and play, or football and gym. The specific details for each score are provided in Table 1.). In numerous research studies involving sentences written in two languages with distinct structures, a common approach has been adopted due to the limited availability of text corpora and labeled data for both languages. Researchers often resort to machine translation to convert sentences from the source language into the target language. Subsequently, they employ semantic similarity models designed for the target language (such as English in the example provided) to gauge the semantic similarity between these translated sentences. Nonetheless, the main weakness of such systems is the error propagation through machine translation and the translation of the source language into the target language may not be done well. One of the primary objectives of the article is to mitigate machine translation errors when determining the similarity between Persian and English sentences. This is achieved through the creation of a dedicated dataset.

In Table 1, a sample of PESTS with their semantic similarity, which is in the range of 0 to 5, can be seen. A score of 5 indicates the highest degree of similarity and 0 indicates the absence of similarity (no similarity).

**Table 1** Sample Persian-English sentence with semantic similarity

| Guide to Similarity Measurement | English Sentences | Persian Sentences | Degree of similarity |
|---|---|---|---|
| The two sentences are completely dissimilar | This is not a request hard to meet. | او درباره دلیل رسیدن به چنین تصمیمی توضیح نداد. (He did not provide an explanation for the reason behind reaching such a decision) | 0 |
| The two sentences are not equivalent, but are on the same topic. (In this case, the verb or one of the capacities of the verb is slightly different. Usually, in one of the two sentences, there is an additional unimportant component that makes little difference in meaning; And it just provides more | A transmission phase of the torch includes Mount Everest, located on the border of Tibetan and Nepal. | وی می‌گوید صعود به بلندترین قله در جهان برایش آرزوئی بوده که سرانجام تحقق یافته است. (He says that climbing the highest peak in the world has been a long-held dream for him, which has finally come true) | 1 |
| information about the verb, or verb capacities.) | | | |
| The two sentences are not equivalent, but share some details. (In this case, the verb or one of the capacities of the verb is different to a great extent, but there is still a semantic similarity between the two components. Usually, there is an additional component in one of the two sentences, which makes a big difference in meaning.) | Six people lost their lives in the attacks and more than one hundred were wounded. | سه نفر از کسانی که جان خود را از دست دادند، ماموران پلیس بودند.<br><br>(Three of those who lost their lives were police officers.) | 2 |
| The two sentences are roughly equivalent, but some important information differs/missing. (In this case, the verb or one of the capacities of the verb is completely different, but the two sentences are similar in at least two important components. limportant information refers to details that readers are interested in and are crucial for comprehending the context.) | The official death toll from this acute and chronic disease of the SARS respiratory has increased in 26 countries to 293 people. | این بیماری مرموز که عوارضی نظیر آنفلونزا دارد تاکنون در سطح جهانی موجب مرگ ۴۴۸ نفر شده است.<br><br>(The mysterious disease, which has symptoms similar to the flu, has so far caused the deaths of 448 people worldwide.) | 3 |
| The two sentences are mostly equivalent, but some unimportant details differ. | Both materials stimulate circulation and help reduce cholesterol levels. | دارچین و عسل گردش خون را تحریک کرده و به کاهش سطح کلسترول کمک می‌کنند.<br><br>(Cinnamon and honey stimulate blood circulation and help reduce cholesterol levels.) | 4 |
| The two sentences are completely equivalent, as they mean the same thing. | Only a decade ago, companies mainly paid attention to research on consumers. | تنها یک دهه است که شرکت‌ها عمدتاً به نتایج تحقیقات بر روی مصرف‌کنندگان توجه می‌کنند.<br><br>(It has only been a decade that companies have mainly focused on research results regarding consumers.) | 5 |

In this paper, the article introduces the Persian-English cross lingual similarity corpus. Subsequently, this corpus is employed to fine-tune various Transformer-based models. Finally, the performance of these models is assessed using a designated test set. The innovations of this article can be listed as follows:
1- Creation of the most extensive Persian-English cross-lingual semantic textual similarity dataset (5375 pairs).
2- Development of a model with the highest correlation to determine the level of semantic similarity between Persian and English sentences.
3- Elimination machine translation step for measuring Persian-English cross lingual semantic textual similarity.

The purpose of this article is to create a semantic similarity dataset between Persian and English. In this context, semantic similarity refers to the semantic distance between two sentences, that is, how similar or different the two sentences are in terms of lexical content and general subject matter. The evaluation outcomes indicate that incorporating the semantic similarity corpus for Persian and English sentences enhances the models' capacity to gauge the degree of similarity, thereby enriching their semantic performance.

In section 2, we first describe the related works in the field of semantic textual similarity. Then, in section 3, we explain how sentence pairs were chosen and annotated. Subsequently, in section 4, we describe the statistics related to the corpus, and finally, we introduce and discuss the semantic textual similarity models that have been developed and tested. The dataset for non-commercial usage has been made publicly available[2].

## 2  Related works

In this section, we briefly introduce elements that are relevant to our work. Semantic similarity is one of the important tasks that various researchers around the world have done, but their main focus has been on English and few corpora have been produced in cross-lingual semantic similarity. To advance the development of cross-lingual semantic similarity, there was a notable effort in 2017, special emphasis was placed on the semantic similarity of English with Arabic, Spanish, and Turkish (Cer *et al.*, 2017). Multiple semantic similarity corpora have been created since 2005, and we highlight a few of them below:

Between 2012 and 2015, several datasets were introduced, and their structures are briefly outlined in Table 5 in the appendix. The dataset from 2012 (Agirre et al., 2012) was drawn from various sources and included 5250 sentence pairs gathered from a wide array of news and video sources in the English language. In 2013, another corpus (Agirre et al., 2013b) comprised 2250 sentence pairs, similar to SemEval 2012 but with variations in the genre of sentence pairs. The STS 2014 dataset, introduced by Agirre et al. (2014), added the Spanish language as a second subtask and consisted of 3750 sentence pairs in English and 804 sentence pairs in Spanish. Furthermore, all datasets published in 2012 and 2013 were employed as training data. STS 2015, as presented by Agirre et al. (2015), introduced three subtasks: English, Spanish, and an interpretable pilot subtask. The interpretable pilot subtask delved into whether systems could elucidate why two sentences were related or unrelated, adding an explanatory layer to the similarity score.

---
[2] https://github.com/mohammadabdous/PESTS

In 2016, (Agirre et al., 2016) introduced a corpus that included cross-lingual test data. The cross-lingual test data was divided into two sets: news and multi-sources. This dataset included 1186 sentence pairs for the English language and 698 sentence pairs for English-Spanish cross-lingual semantic similarity. It's worth noting that for each of these datasets spanning from 2012 to 2016, there was no specific training dataset designated, and it was suggested to utilize older datasets for training purposes. However, in our work, we propose the introduction of distinct sentence pairs as a dedicated training dataset.

Ferrero et al. (Ferrero *et al.*, 2016) provided a dataset to assess the similarity of cross lingual texts that can be very useful in fraud detection. The dataset provided is multilingual, containing texts in French, English, and Spanish languages. It facilitates cross-lingual alignment of information at different levels, such as documents, sentences, and fragments. The dataset comprises both human-generated texts and texts obtained through machine translation. This dataset also includes a variety of documents written by a variety of mid-level to high level writers. This dataset has overcome many of the limitations of previous datasets, one of which is the ability to align only at a certain level (for example, the sentence level).

In the domain of cross-lingual semantic textual similarity, the aim is to estimate the degree of the meaning similarity between two sentences, each in a different language (Hercig and Král, 2021). Semantic similarity assessments typically face fewer difficulties when applied to languages with abundant resources in contrast to those with limited resources. However, for languages lacking appropriate linguistic resources, the task of calculating similarity becomes a significant and formidable challenge.

One potential strategy for addressing this challenge involves employing machine translation-based methods to translate sentences from low-resource languages like Persian into high-resource languages such as English. However, a major drawback of these approaches is the presence of errors in machine translation, and their effectiveness is strongly contingent upon the quality of the translation (Bjerva and Ostling, 2017). In our work, we leverage multilingual Transformers to tackle this issue, eliminating the need for machine translation.

In their study, Tang et al. (Tang *et al.*, 2018) developed a model for low-resource languages such as Spanish, Arabic, Indonesian, and Thai. Using a monolingual semantic similarity model framework, they extended it to multilingual mode. This approach demonstrated that through the utilization of a shared multilingual encoder, each sentence could exhibit distinct embeddings depending on the target language.

At the 2017 Semantic Evaluation Conference (Cer *et al.*, 2017), the main focus was on cross lingual and multilingual semantic textual similarity. In this conference, 17 participants competed in 31 teams and in this conference, the STSBenchMark dataset was presented. Some cross lingual datasets including Arabic-English, Spanish-English and Turkish-English were also introduced and evaluated by the participants.

Important work done in the field of cross lingual semantic similarity is based on cross lingual embeddings (Klementiev, Titov and Bhattarai, 2012)(Zou *et al.*, 2013)(Mikolov, Le and Sutskever, 2013)(Gouws, Bengio and Corrado, 2015)(Ammar *et al.*, 2016).

Chidambaram et al. (Chidambaram *et al.*, 2018) have generated cross lingual vector space using a model based on dual encoder. Their goal is to train a model that produces the maximum similarity between sentence pairs in a paraphrase corpus. The resulting embeddings are improved through the incorporation of monolingual datasets and concurrent multitasking training. They use the common vector space obtained in many tasks and have a comparative advantage compared to other tasks. At the core of their proposed method is modeling the tasks which is based on ranking sentence pairs using dual encoders and producing cross lingual embeddings by leveraging machine translations. In shared encoder architecture, there are three Transformers, each with feed forward sublayers and multi head attention. The Transformer's output consists of a variable-length sequence, and by averaging these sequences, sentence embeddings are generated. Subsequently, the embeddings produced in the feed-forward layers are employed to fine-tune each specific task.

Conneau et al. (Conneau *et al.*, 2018) have developed a cross lingual dataset called XNLI. Because data collection in all languages is a costly process, the interest in Cross-Lingual Language Understanding and transmission in low-resource languages has increased. In this paper, a test dataset for Cross-Lingual Language Understanding has been developed and the test datasets have been expanded to 15 languages, including low resource languages such as Swahili and Urdu. Labels in the generated corpus are premise and hypothesis. To demonstrate the usefulness of the dataset, they conducted experiments on various tasks, including machine translation, multilingual bag of words, and LSTM encoder.

Conneau et al. (Conneau and Lample, 2019) have proposed a cross lingual model called XLM. They used two methods to learn cross lingual models. The first method is unsupervised, which relies only on monolingual data, and the second method is supervised, which uses a paraphrase corpus with the aim of modeling the target language. Their proposed method performs best on supervised and unsupervised machine translation tasks and XNLI. Table 5 in the appendix summarizes the generated semantic textual similarity datasets.

The masked language model has a similar purpose as introduced by Devlin (Devlin et al., 2018), except that in this model we are faced with a continuous flow of sentence pairs. In the machine translation language model, like the masked language model, pairs of parallel sentences are given to the machine. To predict an English word, this model can look at both English and French translations, and it seeks to align English and French embeddings. XLM uses the Byte-Pair Encoding (BPE) method (Shibata et al, 1999) and the Bert language learning mechanism to learn the relation between words in different languages. Byte-Pair Encoding is a data compression method that consistently replaces the most frequently repeated character pairs (essentially bytes) in a particular dataset with a non-event symbol in the text. In each iteration, the algorithm finds the most frequent pairs of characters and merges them to create a new symbol. In the XLM model, instead of using the word or characters as input to the model, it uses Byte-Pair Encoding, which divides the input into the most common sub-word in all languages, thus increasing common cross-lingual vocabulary. The XLM model enhances Bert's architecture in two ways:

In the Bert model, each train instance consists of one language, while in the XLM model, each train instance consists of two languages. As in the prediction of masked words in Bert's model, this model uses the context of the source sentence to predict the masked words of the destination sentence. This model also receives the language identifier and word order in each language separately as a position encoder. This new metadata helps the model learn the relation between related words in different languages. In our work, we employ Bert and XLM models to assess the performance of our dataset.

In this article, a semantic similarity between Persian and English sentences was created. In our dataset, we sought inspiration from the advantages of older datasets, particularly the main sources they utilized, while simultaneously addressing their disadvantages, such as the reliance on machine translation. In the following, we will explain how to produce the dataset.

## 3 Corpus production process

As previously mentioned, the process of constructing the cross-lingual corpus begins with the creation of a Persian-Persian semantic similarity corpus. This initial step is undertaken because there is greater ease in finding expert linguists who possess the ability to discern semantic distinctions among Persian sentences. Simultaneously identifying linguistic experts proficient in both English and Persian presents a more formidable challenge.

In order to produce the Persian-Persian corpus, first 20,000 pairs of suitable Persian sentences were selected and the semantic scores between them were presented by the annotators; after that, a part of the whole corpus was used to make a Persian-English corpus. In the following, the details of corpus production will be described.

Constructing our datasets, collecting pairs of natural sentences with varying scores of semantic similarities was itself a challenge. If we consider a pair of sentences at random, the majority of them will be completely unrelated, and only a very small fragment will show some kind of semantic equivalent.

### 3.1 Sentence sources

Identifying similar sentence pairs presents a significant challenge. Neglecting this critical aspect during corpus production can often result in the majority of sentence-based corpora having a degree of similarity 0 or 1, which, in turn, diminishes the corpus's quality for applications involving machine learning and real-world scenarios.

To address this challenge, we adopted a strategy involving multiple textual sources and various text corpora. In the process of constructing our corpus, we made use of the Persika corpus (Eghbalzadeh et al., 2012) and the Hamshahri corpus (AleAhmad et al., 2009), both of which contain Farsi news data. However, it's essential to emphasize that the data sourced from these repositories is not pristine; hence, thorough preprocessing is necessary to cleanse and prepare it for subsequent analysis and practical usage.

In our preprocessing strategy sentence pairs ranging from 7 to 25 words in length were extracted from these corpora. Sentences must carry semantic meaning and exhibit distinguishable semantic differences. For instance, a sentence like "او آمد" ("he comes") lacks the necessary semantic complexity and is not suitable for labeling. Additionally, sentences with very short lengths are not ideal candidates for inclusion in the final corpus. Furthermore, sentences with an excessive number of words can present challenges for annotators, as they may contain too many details, making it difficult to discern the main differences between the two sentences. As noted in the literature, the average sentence length in parallel corpora for Persian sentences is approximately 11 words per sentence (Kashefi, 2018). Many Persian corpora fall within the range of 7 to 25 words in length, including the Persian Discourse Treebank and Coreference Corpus (Mirzaei and Safari, 2018), Hamshahri: A Standard Persian Text Collection (AleAhmad et al., 2009), and TEP: Tehran English-Persian Parallel Corpus (Pilevar et al., 2011). These averages are indicative of the suitable sentence lengths for inclusion in our corpus. It's worth noting that there are additional preprocessing steps for selecting sentence pairs, which will be detailed below.

## 3.2 Sentence preprocessing

In extensive text corpora like Ferdowsi, Persica (Eghbalzadeh et al., 2012), Wikipedia, and similar sources, a notable issue arises from the absence of proper pre-processing of the sentences contained within them. This lack of preprocessing can lead to various problems and challenges in working with these corpora. Instances of sentences containing only two words are present in the corpus, and spelling errors frequently occur within sentences, necessitating correction or removal. Additionally, normalization is crucial, as characters such as 'ی' and 'ي' should be unified to ensure consistency. Without proper normalization, models may generate various vectors for sentences conveying the same meaning. For instance, the sentences '.علی رضا از بوستان فدک به خانه آمد' and ' علي رضا از بوستان فدك به خانة امد. ' should yield identical vectors after preprocessing; failure to do so results in disparate vector representations. To make these corpora practically useful, it is imperative to thoroughly examine the sentences and texts within them and address their shortcomings. In general, preprocessing plays a significant role in influencing the output quality of transformers and deep neural networks such as Fast text and Glove (Rahimi & Homayounpour, 2023). In the following, we will elaborate on the diverse preprocessing steps that were undertaken to achieve this goal:

1. From the text corpus, sentences were selected that have a number of words between 7 and 25 (As mentioned in section 3.1). This selection criterion helped ensure that the sentences in the corpus met the desired length requirements for subsequent analysis and processing.

2. To uphold the desired quality and language consistency of the corpus, sentences containing elements from other languages, even if they consisted of just one or more words in a different language, were methodically eliminated. Language detection tools like "[langDetect python](#)"[3] were employed to identify and flag sentences with content in languages other than the target languages (Persian). This approach helped ensure that the corpus remained focused on the intended languages and maintained its quality standards.

3. The sentences within the corpus underwent a comprehensive spell-checking process to rectify any misspelled words. This spell-checking procedure involved the use of two distinct tools:

    1) [Python-negar](#) library[4] for the Persian language.
    2) [Pyspellchecker](#) library[5] for the English language.
    3) A linguistic expert with a Ph.D. in linguistics also reviewed and validated the sentences. This multi-step approach was employed to ensure the accuracy and correctness of the sentences in both Persian and English. Human experts have ensured that some sentence pairs selected from the corpora exhibit a semantic similarity exceeding a degree of 2, to ensure a sufficient number of sentences within each degree of similarity interval.
    4) The normalizer tool ([hazm](#) package[6]) has been employed to analyze the characteristics of the words in the sentences.
    5) If a sentence includes a quotation, the section following the ":" (referred to as the "quote part") is chosen as the primary sentence. In the Persian sentence, " وی گفت: تمامی کشورها باید در مسیر صلح و دوستی گام بردارند." ("All countries should take a step in the path of peace and friendship," he said), only the segment " تمامی کشورها باید در مسیر صلح و دوستی گام

---

[3] https://pypi.org/project/langdetect

[4] https://pypi.org/project/python-negar

[5] https://pypi.org/project/pyspellchecker

[6] https://pypi.org/project/hazm

بردارند." ("All countries should take a step in the path of peace and friendship") is retained.
  6) Sentence length discrepancies of up to 5 words are acceptable.

## 3.3 Selection of sentence pairs

Following sentence pre-processing, sentence pairs were selected by an independent human expert (not part of the annotators) from the corpus. This aimed to ensure that the resulting corpus exhibited a desirable distribution of similarity scores, encompassing all possible scores from 0 to 5, and containing appropriately matched sentence pairs. When sentence pairs were randomly chosen from the pool of pre-processed sentences, over 99% of them yielded degree of similarity of either 0 or 1. In reality, many of these sentence pairs exhibited limited semantic similarity, and their distribution of similarity scores was far from uniform.

## 3.4 Annotation process

In the production process of large corpora, diverse annotation systems are employed. The utilization of these annotation systems serves to enhance the quality and speed of corpus production. For the creation of the semantic textual similarity corpus, we employed a web-based system comprising React for the front end and Django for the back end, with PostgreSQL as the database which we will describe in the following.

The annotating system for sentence pairs consists of two main components. In the first part, each annotator can view sentence pairs related to them using their provided user ID and assign a score ranging from -1 to 5 to each pair of sentences. This scoring mechanism accounts for the possibility that, despite prior reviews and preprocessing, the sentence pairs may still contain misspellings or other semantic issues. If an annotator identifies such issues, they assign a score of -1. Finally, another expert reviews the scores of -1 and if the sentence pair cannot be corrected, it will be removed from the dataset; otherwise, the sentence pair is corrected and resent for annotation with a particular emphasis placed on strict adherence to the labeling guidelines. The labeling guideline has instructions for various challenging cases. In the different sessions we had with the annotators, many discussions have taken place regarding challenging sentences, as outlined in the previous section.

In some cases, discrepancies may arise in the assigned scores. The differences between annotators usually occur in scores that are close, such as between 4 and 5. This variation is due to their diverse opinions about the semantic similarity of certain words in two sentences. For instance, one annotator might consider two words in two sentences as having the same meaning, but another annotator may discover some differences in meaning.

To illustrate, let's consider a similarity score of 3. In such cases, the two sentences can be nearly synonymous, but there are notable differences or omissions in one of the sentences. These differences might involve significant variations in verbs or the capacities of the verbs, yet a fundamental semantic similarity still exists between the components. Typically, one of the two sentences includes an additional element that significantly impacts the overall meaning. Occasionally, when different annotators assess the sentences, one may assign a score of three while the other assigns a score of four based on their personal judgment. For instance, in the sentences "یک سگ از روی نرده‌های پارک می‌پرد" and (.A dog is playing in the park) یک سگ در پارک بازی می‌کند.

(A dog jumps over the fence in the park.)" one annotator might give a score of three, and the other might give a score of four based on their assessments.

In the second part of the system, the production of the corpus is managed. By entering this system, the admin user can see the sentence pairs that have been annotated by all annotators so far according to different filters such as hours, number of annotator users, average score of three annotators, and so on. Users can also use this system to modify or edit the score of sentence pairs that have not been annotated yet by themselves before. For annotating each pair by experts, the instruction that is described in Table 1 is used.

### 3.5 Experts specification

To assign scores to sentence pairs, three linguists with a Ph.D. degree in linguistics and highly experienced in the field of linguistic and textual data analysis have been used. All annotators follow the same basis and guidelines for scoring. The correlation between annotators is important and causes the corpus to be more accurate. Table 2 presents the correlation levels among annotators based on the scores they assigned to the sentence pairs.

**Table 2** The degree of correlation between annotators' scores

| Title | measurement |
|---|---|
| Annotator1 & annotator 2 (Pearson correlation) | 90.32% |
| Annotator 1 & average scores of annotators 2 and 3 (Pearson correlation) | 92.66% |
| Annotator1 & annotator 3 (Pearson correlation) | 90.80% |
| Annotator 2 & average scores of annotators 1 and 3 (Pearson correlation) | 92.86% |
| Annotator2 & annotator 3 (Pearson correlation) | 92.05% |
| Annotator 3 & average scores of annotators 1 and 2 (Pearson correlation) | 93.21% |
| Percent of cases that 3 annotator's score is same (0 or 1 or 2 or 3 or 4 or 5) | 43.40 % |
| Average variance of annotator's label | 0.13 |
| Average standard deviation of annotator's label | 0.27 |
| The percentage of cases where taggers 1 and 2 assigned identical scores. | 60.10% |
| The percentage of cases where taggers 2 and 3 assigned identical scores. | 62.40% |
| The percentage of cases where taggers 1 and 3 assigned identical scores. | 63.90% |

As indicated in Table 2, the Pearson correlation coefficient between the annotations exceeds 90%, which is considered a highly favorable level of agreement.

### 3.6 Review of sentence pairs by an expert

Despite employing the same scoring criteria, annotators frequently assign varying scores to a single pair, and several factors contribute to this variance. Firstly, an annotator may make scoring errors due to factors such as fatigue, lack of concentration, or the speed at which they work. The second factor stems from the ambiguity in defining semantic similarity, often relying on individual opinions to gauge the extent of similarity. Lastly, our scoring system uses distinct boundaries (ranging from 0 to 5), while the actual degree of similarity may fall between two whole numbers. For instance, if the true score hovers around 2.5, one annotator may assign a score of 2, while another may give it a 3.

In the final corpus, the ultimate semantic similarity score for a sentence pair is derived from the average score assigned by three annotators. Consequently, when score discrepancies result from individual errors, they should be rectified. However, if the variations are related to the two other reasons (unclear similarity concepts or fixed score boundaries), they are acceptable as long as the difference in scores remains within a one-unit range. Therefore, significant score disparities (i.e., greater than one unit) often indicate a scoring error and necessitate further review. If the difference between the scores of two annotators exceeds 2, it is advisable to seek the input of an expert to understand the underlying reasons for this discrepancy.

### 3.7 Creating a Persian-English corpus

To create a dataset that accurately represents a uniform distribution of similarity score ranges, a subset comprising 5,374 sentence pairs was selected and extracted from a larger pool of over 20,000 scored Persian sentence pairs. In this process, the initial sentence in each pair was translated from Persian to English by proficient linguists, effectively replacing the Persian sentences. As a result, a corpus was formed in which the first component of the sentence pairs is in English, the second component is in Persian, and the degree of similarity between the Persian and English sentences is denoted in the third column.

### 3.8 Statistics of corpus

Within the produced corpus, a total of 5,374 sentence pairs underwent evaluation by three annotators. Subsequently, a semantic similarity model was developed using the training set, comprising 90% of these 5,374 sentence pairs. The statistics detailing this generated corpus are provided in Table 3, and these statistics can be applied to various natural language processing tasks.

**Table 3** Statistics of the corpus

| | Train (80%) | Dev (10%) | Test (10%) | All (100%) |
|---|---|---|---|---|
| Number of Pairs | | | | |
| all score | 4298 | 538 | 538 | 5375 |
| scores between 0 to 1 | 920 | 130 | 131 | 1181 |
| scores between 1 to 2 | 488 | 60 | 60 | 608 |
| scores between 2 to 3 | 1087 | 136 | 136 | 1359 |
| scores between 3 to 4 | 640 | 80 | 80 | 800 |
| scores between 4 to 5 | 1163 | 132 | 131 | 1427 |
| Average number of words in a persian sentence | 14.17 | 13.88 | 14.27 | 14.10 |
| Average number of words in an English sentence | 14.00 | 13.84 | 14.03 | 13.97 |

The data is divided into training (80%), testing (10%), and development (10%) sets with random sampling in each section to maintain the primary distribution of each score.

## 4 Results and discussion

In this paper, we have performed experiments on the generated corpus, employing a range of Transformer-based language models. Our objective was to assess and compare the performance of these models in tasks associated with quantifying semantic textual similarity. Transformers have been developed to address the sequencing problem in neural networks. They demonstrate exceptional proficiency in handling sequences of data, such as the words in a sentence. They meticulously scrutinize each element of the input sequence before generating the output, and this careful processing persists until the entire sequence has been thoroughly analyzed. Transformers consist of encoders that play a crucial role in capturing the meanings and concepts within the data and transforming them into semantic vectors. Some of these Transformers are multilingual, such as Multilingual BERT (mBERT) and XLM (Cross-lingual Language Model). These models are trained across various languages and can effectively provide vector representations for different languages. Notably, the performance of these Transformers can be further enhanced by fine-tuning them using cross-lingual corpora. It's important to mention that during the fine-tuning process, the weights of the last layer in the Transformers are updated based on the training data. This strategy is adopted for several reasons. Firstly, the earlier layers of the Transformer model, having been pre-trained on large-scale corpora, often encode more generic linguistic features such as syntax and basic linguistic patterns. These features are generally transferable across different tasks and languages and thus do not require extensive modification. On the other hand, the later layers of the Transformer tend to capture more task-specific features and nuances, such as semantics and higher-level linguistic structures. In the case of our specific task, which revolves around cross-lingual semantic textual similarity, these semantic features are of paramount importance. By selectively updating the parameters of the last layers while keeping the initial layers fixed, we can leverage the pre-trained knowledge of the earlier layers to facilitate the learning of task-specific representations in the final layers (Jawahar et al., 2019). The weights of the last layer in a Transformer model refer to the parameters associated with the final self-attention and feedforward neural network layers in the model architecture. These weights include the learnable parameters such as the weight matrices and bias vectors associated with the final attention mechanism and the subsequent feedforward neural network layers. Freezing the initial layer during the fine-tuning step also results in a significant reduction in computational and storage costs.

In these experiments, we fine-tuned the models by utilizing the training data from the Persian-English semantic similarity corpus. The results demonstrated an enhancement in their performance when compared to the non-fine-tuned mode, as illustrated in Table 4.

During the fine-tuning process in our experiments, we conducted training over four epochs with a batch size of 32. The loss function employed was the cosine similarity loss function.

The cosine similarity criterion has been employed to measure the semantic similarity between the two sentences. The criterion of cosine similarity between two vectors is one of the most widely used criteria in measuring semantic similarity between sentences.

### 4.1 evaluation measure

Pearson correlation coefficient (Benesty *et al.*, 2009) and Spearman (SPEARMAN, 1910) are used to evaluate the output of semantic textual similarity systems. The purpose is to calculate the correlation between the score of similarity detected by the system and its true score of similarity. How to calculate Pearson correlation coefficient according to Equation 1:

$$\Gamma_{XY} = \frac{\sum_{i=1}^{n}(x_i-\bar{x})(y_i-\bar{y})}{\sqrt{\sum_{i=1}^{n}(x_i-\bar{x})^2}\sqrt{\sum_{i=1}^{n}(y_i-\bar{y})^2}} \quad (1)$$

In the above formula $x_i$ indicates first (or predicted) score and $y_i$ indicates the second (or gold) score. $\bar{x}$ indicate the average of first (or predicted) scores and $\bar{y}$ indicate the average of second (or gold) scores. Predicted or gold score is used in the testing step.

When the Pearson correlation coefficient approaches one, it indicates that the model's performance is highly accurate. Below, we present the results obtained from applying the model to the test data for measuring semantic textual similarity between English and Persian languages.

### 4.2 experiments

To assess the performance of the developed model, we conducted tests and evaluations using cross-lingual semantic textual similarity test data. We employed the cosine similarity metric to compute the semantic similarity score between sentence vectors, and the Pearson and Spearman correlation coefficients were used to gauge the correlation between the model's scores and the gold standard scores.

The results, presented in Table 4, highlight that by utilizing the generated corpus, the Transformer-based models exhibit an improved ability to measure semantic similarity between English and Persian. For instance, the "paraphrase-xlm-r-multilingual-v1" model (Reimers and Gurevych, 2020) achieved a Pearson correlation of 85.87 in the non-fine-tuned mode and an impressive Pearson correlation of 95.62 in the fine-tuned mode. This fine-tuning resulted in an approximately ten percent increase in the correlation rate when compared to the non-fine-tuned mode, underscoring the efficacy of the generated corpus.

It's worth noting that throughout all experiments, the test data for evaluation and the training data for fine-tuning remained consistent.

**Table 4** Results of the implementation of multilingual Transformers-based models for cross lingual semantic textual similarity

| Models[7] | Without fine-tuning | | Fine-tuned model | |
|---|---|---|---|---|
| | Pearson | Spearman | Pearson | Spearman |
| Xlm-roberta base (Conneau *et al.*, 2019) | 23.80 | 28.99 | 89.48 | 90.40 |
| paraphrase-xlm-r-multilingual-v1 (Reimers and Gurevych, 2020) | **85.87** | **85.91** | **95.62** | **95.17** |
| bert-base-multilingual-cased (Reimers and Gurevych, 2019) | 47.00 | 45.47 | 91.88 | 91.55 |
| distilbert-base-multilingual-cased (Sanh et al., 2019) | 45.56 | 45.18 | 89.51 | 89.08 |
| stsb-xlm-r-multilingual (Reimers and Gurevych, 2019) | 78.37 | 76.57 | 94.43 | 94.02 |
| xlm-r-100langs-bert-base-nli-stsb-mean-tokens (Reimers and Gurevych, 2019) | 78.37 | 76.57 | 94.40 | 94.03 |
| Twitter-xlm-roberta base (Barbieri et al, 2021) | 27.94 | 28.06 | 90.99 | 90.28 |

## 5 Conclusion

In today's era, as textual resources continue to grow in various languages, there is a growing demand for models that can effectively comprehend multiple languages simultaneously. One crucial task in natural language processing is the comprehension of sentence or phrase meanings, often referred to as semantic textual similarity. Semantic textual similarity represents a significant area of research in natural language processing, particularly in the context of cross-lingual applications.

---
[7] https://huggingface.co/models

In this study, we have developed a corpus dedicated to cross-lingual semantic textual similarity. To create this corpus, we initially generated a Persian-Persian dataset and then had proficient linguists translate the first part of each sentence pair into English.

Our experiments conducted in this research underscore the significance of producing such a corpus. It serves the dual purpose of enhancing the performance of models for semantic textual similarity tasks involving Persian and English languages and enabling the evaluation and comparison of these models using the corpus's test data.

Furthermore, the models we've created can be used in many applications such as question-answering systems, fraud detection, machine translation, and information retrieval in both Persian and English contexts.

**Data Availability**
The annotated data is available on GitHub(https://github.com/mohammadabdous/PESTS)

**Funding**
This research received no grant from any funding agency.

**Author Contributions**
Prepare tagging manual and dataset: M.A and P.P; Programming and Modelling: M.A and P.P; Supervision: B.M.

**Corresponding Author**
Behrouz Minaei Bidgoli is corresponding author of this article.

**Declaration**
**Conflict of interest:** The authors have no competing interests to declare.

**References**

**Agirre, E., Cer, D., Diab, M. and Gonzalez-Agirre, A.,** 2012. Semeval-2012 task 6: A pilot on semantic textual similarity. In * SEM 2012: The First Joint Conference on Lexical and Computational Semantics–Volume 1: Proceedings of the main conference and the shared task, and Volume 2: Proceedings of the Sixth International Workshop on Semantic Evaluation (SemEval 2012) (pp. 385-393).

**Agirre, E., Cer, D., Diab, M., Gonzalez-Agirre, A. and Guo, W.,** (2013A) * SEM 2013 shared task: Semantic textual similarity. In *Second joint conference on lexical and computational semantics (* SEM), volume 1: proceedings of the Main conference and the shared task: semantic textual similarity* (pp. 32-43).

**Agirre, E., Banea, C., Cardie, C., Cer, D.M., Diab, M.T., Gonzalez-Agirre, A., Guo, W., Mihalcea, R., Rigau, G. and Wiebe, J.,** 2014, August. SemEval-2014 Task 10: Multilingual Semantic Textual Similarity. In SemEval@ COLING (pp. 81-91).

**Agirre, E., Banea, C., Cardie, C., Cer, D., Diab, M., Gonzalez-Agirre, A., Guo, W., Lopez-Gazpio, I., Maritxalar, M., Mihalcea, R. and Rigau, G.,** 2015, June. Semeval-2015 task 2: Semantic textual similarity, english, spanish and pilot on interpretability. In *Proceedings of the 9th international workshop on semantic evaluation (SemEval 2015)* (pp. 252-263).

**Agirre, E., Banea, C., Cer, D., Diab, M., Gonzalez Agirre, A., Mihalcea, R., Rigau Claramunt, G. and Wiebe, J.,** 2016. Semeval-2016 task 1: Semantic textual similarity, monolingual and cross-lingual evaluation. In SemEval-2016. 10th International Workshop on Semantic Evaluation; 2016 Jun 16-17; San Diego, CA. Stroudsburg (PA): ACL; 2016. p. 497-511.. ACL (Association for Computational Linguistics).

**Al-Anzi, F.S. and AbuZeina, D.,** 2017. Toward an enhanced Arabic text classification using cosine similarity and Latent Semantic Indexing. Journal of King Saud University-Computer and Information Sciences, 29(2), pp.189-195.

**Aliguliyev, R.M.,** 2009. A new sentence similarity measure and sentence based extractive technique for automatic text summarization. Expert Systems with Applications, 36(4), pp.7764-7772.

**Alzahrani, S.M., Salim, N. and Abraham, A.,** 2011. Understanding plagiarism linguistic patterns, textual features, and detection methods. *IEEE Transactions on Systems, Man, and Cybernetics, Part C (Applications and Reviews)*, *42*(2), pp.133-149.

**Ammar, W., Mulcaire, G., Tsvetkov, Y., Lample, G., Dyer, C. and Smith, N.A.,** 2016. Massively multilingual word embeddings. arXiv preprint arXiv:1602.01925.

**Barbieri, F., Anke, L.E. and Camacho-Collados, J.,** 2021. Xlm-t: Multilingual language models in twitter for sentiment analysis and beyond. *arXiv preprint arXiv:2104.12250*.

**Benesty, J., Chen, J., Huang, Y. and Cohen, I.,** 2009. Pearson correlation coefficient. In Noise reduction in speech processing (pp. 1-4). Springer, Berlin, Heidelberg.

**Bjerva, J. and Östling, R.,** 2017. Cross-lingual learning of semantic textual similarity with multilingual word representations. In 21st Nordic Conference on Computational Linguistics, NoDaLiDa, Gothenburg, Sweden, 22-24 May, 2017 (pp. 211-215). Linköping University Electronic Press.

**De Boni, M. and Manandhar, S.,** 2003. The Use of Sentence Similarity as a Semantic Relevance Metric for Question Answering. In New Directions in Question Answering (pp. 138-144).

**Brychcín, T.,** 2020. Linear transformations for cross-lingual semantic textual similarity. Knowledge-Based Systems, 187, p.104819.

**Cer, D., Diab, M., Agirre, E., Lopez-Gazpio, I. and Specia, L.,** 2017. Semeval-2017 task 1: Semantic textual similarity-multilingual and cross-lingual focused evaluation. arXiv preprint arXiv:1708.00055.

**Chidambaram, M., Yang, Y., Cer, D., Yuan, S., Sung, Y.H.,** Strope, B. and Kurzweil, R., 2018. Learning cross-lingual sentence representations via a multi-task dual-encoder model. arXiv preprint arXiv:1810.12836.

**Conneau, A., Lample, G., Rinott, R., Williams, A., Bowman, S.R., Schwenk, H. and Stoyanov, V.,** 2018. XNLI: Evaluating cross-lingual sentence representations. arXiv preprint arXiv:1809.05053.

**Conneau, A., Khandelwal, K., Goyal, N., Chaudhary, V., Wenzek, G., Guzmán, F., Grave, E., Ott, M., Zettlemoyer, L. and Stoyanov, V.,** 2019. Unsupervised cross-lingual representation learning at scale. arXiv preprint arXiv:1911.02116.

**Conneau, A. and Lample, G.,** 2019. Cross-lingual language model pretraining. Advances in neural information processing systems, 32.

**Devlin, J., Chang, M.W., Lee, K. and Toutanova, K.,** 2018. Bert: Pre-training of deep bidirectional Transformers for language understanding. arXiv preprint arXiv:1810.04805.

**Eghbalzadeh, H., Hosseini, B., Khadivi, S. and Khodabakhsh, A.,** 2012, November. Persica: A Persian corpus for multi-purpose text mining and Natural language processing. In 6th International Symposium on Telecommunications (IST) (pp. 1207-1214). IEEE.

**Ferrero, J., Agnes, F., Besacier, L. and Schwab, D.,** 2016, May. A multilingual, multi-style and multi-granularity dataset for cross-language textual similarity detection. In 10th edition of the Language Resources and Evaluation Conference..

**Gouws, S., Bengio, Y. and Corrado, G.,** 2015, June. Bilbowa: Fast bilingual distributed representations without word alignments. In International Conference on Machine Learning (pp. 748-756). PMLR.

**Hercig, T. and Král, P.**, 2021, September. Evaluation Datasets for Cross-lingual Semantic Textual Similarity. In *Proceedings of the International Conference on Recent Advances in Natural Language Processing (RANLP 2021)* (pp. 524-529)

**Jawahar, G., Sagot, B. and Seddah, D.,** 2019, July. What does BERT learn about the structure of language?. In ACL 2019-57th Annual Meeting of the Association for Computational Linguistics

**Kashefi, O.**, 2018. MIZAN: a large persian-english parallel corpus. arXiv preprint arXiv:1801.02107.

**Klementiev, A., Titov, I. and Bhattarai, B.,** 2012, December. Inducing crosslingual distributed representations of words. In Proceedings of COLING 2012 (pp. 1459-1474).

**Lee, M. D., Pincombe, B. and Welsh, M.,** 2005 'An empirical evaluation of models of text document similarity', in Proceedings of the annual meeting of the cognitive science society.

**Li, Y., McLean, D., Bandar, Z.A., O'shea, J.D. and Crockett, K.,** 2006. Sentence similarity based on semantic nets and corpus statistics. IEEE transactions on knowledge and data engineering, 18(8), pp.1138-1150.

**Majumder, G., Pakray, P., Gelbukh, A. and Pinto, D.,** 2016. Semantic textual similarity methods, tools, and applications: A survey. Computación y Sistemas, 20(4), pp.647-665.

**Manjula, D. and Geetha, T.V.,** 2004. Semantic search engine. Journal of Information & Knowledge Management, 3(01), pp.107-117.

**Marelli, M., Menini, S., Baroni, M., Bentivogli, L., Bernardi, R. and Zamparelli, R.,** 2014, May. A SICK cure for the evaluation of compositional distributional semantic models. In Proceedings of the Ninth International Conference on Language Resources and Evaluation (LREC'14) (pp. 216-223).

**Mayank, M.,** 2020. Intrinsic analysis for dual word embedding space models. arXiv preprint arXiv:2012.00728.

**Mikolov, T., Le, Q.V. and Sutskever, I.,** 2013. Exploiting Similarities among Languages for Machine Translation.

**Mirzaei, A. and Safari, P.,** 2018, May. Persian discourse treebank and coreference corpus. In Proceedings of the eleventh international conference on language resources and evaluation (Irec 2018).

**Pilevar, M.T., Faili, H. and Pilevar, A.H.,** 2011, February. Tep: Tehran english-persian parallel corpus. In International Conference on Intelligent Text Processing and Computational Linguistics (pp. 68-79). Berlin, Heidelberg: Springer Berlin Heidelberg.

**Rahimi, Z., & Homayounpour, M. M.** (2023). The impact of preprocessing on word embedding quality: A comparative study. *Language Resources and Evaluation*, *57*(1), 257-291.

**Reimers, N. and Gurevych, I.,** 2019, November. Sentence-BERT: Sentence Embeddings using Siamese BERT-Networks. In Proceedings of the 2019 Conference on Empirical Methods in Natural Language Processing and the 9th International Joint Conference on Natural Language Processing (EMNLP-IJCNLP) (pp. 3982-3992).

**Reimers, N. and Gurevych, I.,** 2020, November. Making Monolingual Sentence Embeddings Multilingual using Knowledge Distillation. In Proceedings of the 2020 Conference on Empirical Methods in Natural Language Processing (EMNLP) (pp. 4512-4525).

**Sanh, V., Debut, L., Chaumond, J. and Wolf, T.**, 2019. DistilBERT, a distilled version of BERT: smaller, faster, cheaper and lighter. *arXiv preprint arXiv:1910.01108*.

**Shibata, Y., Kida, T., Fukamachi, S., Takeda, M., Shinohara, A., Shinohara, T. and Arikawa, S.,** 1999. Byte Pair encoding: A text compression scheme that accelerates pattern matching.

**Spearman, C.,** 1910. Correlation calculated from faulty data. British journal of psychology, 3(3), p.271.

**Tang, X., Cheng, S., Do, L., Min, Z., Ji, F., Yu, H., Zhang, J. and Chen, H.,** 2018. Improving multilingual semantic textual similarity with shared sentence encoder for low-resource languages. arXiv preprint arXiv:1810.08740.

**Žižka, J. and Dařena, F.,** 2010, September. Automatic sentiment analysis using the textual pattern content similarity in natural language. In International Conference on Text, Speech and Dialogue (pp. 224-231). Springer, Berlin, Heidelberg.

**Zou, W.Y., Socher, R., Cer, D. and Manning, C.D.,** 2013, October. Bilingual word embeddings for phrase-based machine translation. In Proceedings of the 2013 conference on empirical methods in natural language processing (pp. 1393-1398).

**Appendix**

Table 5 Past semantic textual similarity datasets

| Corpus | Language | Dataset | Pairs |
|---|---|---|---|
| SemEval -2012 Task 6 (Agirre *et al.*, 2012) | English | MSRpar | 1500 |
| | | MSRvid | 1500 |
| | | OnWN | 750 |
| | | SMTnews | 750 |

| | | SMTeuroparl | 750 |
|---|---|---|---|
| | | **SUM** | **5250** |
| SEM 2013 shared task (Agirre *et al.*, 2013a) | English | HDL | 750 |
| | | FNWN | 189 |
| | | OnWN | 561 |
| | | SMT | 750 |
| | | **SUM** | **2250** |
| SemEval -2014 Task 10 (Agirre *et al.*, 2014) | English | HDL | 750 |
| | | OnWN | 750 |
| | | Deft-forum | 450 |
| | | Deft-news | 300 |
| | | Images | 750 |
| | | Tweet-news | 750 |
| | | **SUM** | **3750** |
| | Spanish | Wikipedia | 324 |
| | | News | 480 |
| | | **SUM** | **804** |
| SemEval -2015 task 2 (Agirre *et al.*, 2015) | English | HDL | 750 |
| | | Images | 750 |
| | | Answer-student | 750 |
| | | Answer-forum | 375 |
| | | Belief | 375 |
| | | **SUM** | **3000** |
| | Spanish | Wikipedia | 251 |
| | | News | 500 |
| | | **SUM** | **751** |
| SemEval -2016 task 1 (Agirre *et al.*, 2016) | English | HDL | 249 |
| | | Plagiarism | 230 |
| | | Postediting | 244 |

| | | Ans.-Ans. | 254 |
| --- | --- | --- | --- |
| | | Quest.-Quest. | 209 |
| | | **SUM** | **1186** |
| | Spanish-English | Trial | 103 |
| | | News | 301 |
| | | Multi-source | 294 |
| | | **SUM** | **698** |
| SICK (Marco Marelli *et al.*, 2014) | English | SemEval 2012 MSRvid ImageFlick | 9840 |
| SemEval -2017 task1 (Cer *et al.*, 2017) | English | evaluation | 250 |
| | | Trial | 23 |
| | | **SUM** | **273** |
| | Spanish | evaluation | 250 |
| | | Trial | 23 |
| | | **SUM** | **273** |
| | Arabic | evaluation | 250 |
| | | Trial | 23 |
| | | MSRpar | 510 |
| | | MSRvid | 368 |
| | | SMTeuroparl | 203 |
| | | **SUM** | **1354** |
| | Spanish-English | evaluation | 500 |
| | | Trial | 23 |
| | | MT | 1000 |
| | | **SUM** | **1523** |
| | Arabic-English | evaluation | 250 |
| | | Trial | 23 |

|  |  | MSRpar | 1020 |
|---|---|---|---|
|  |  | MSRvid | 736 |
|  |  | SMTeuroparl | 406 |
|  |  | **SUM** | **2435** |
|  | Turkish-English | evaluation | 250 |